\documentclass[11pt,letterpaper]{article}
\usepackage{emnlp2016}
\usepackage{times}
\usepackage{url}
\usepackage{latexsym}
\usepackage{graphicx}
\usepackage{amsmath}
\usepackage{enumerate}
\usepackage{amsfonts}
\usepackage{array}
\usepackage{subfigure}
\usepackage{amssymb}
\usepackage{multirow}
\usepackage{booktabs}
\usepackage{footmisc}

\emnlpfinalcopy

\title{Knowledge Representation via Joint Learning of Sequential Text and Knowledge Graphs}

\author{Jiawei Wu$^1$, Ruobing Xie$^2$, Zhiyuan Liu$^{2}\thanks{$^*$Corresponding author: Z. Liu (liuzy@tsinghua.edu.cn).}$, Maosong Sun$^2$\\
  $^1$School of Medicine, Tsinghua University, Beijing, China\\
  $^2$State Key Laboratory of Intelligent Technology and Systems\\
  Tsinghua National Laboratory for Information Science and Technology\\
  Department of Computer Science and Technology, Tsinghua University, Beijing, China}

\date{}

\begin{document}

\maketitle

\begin{abstract}

Textual information is considered as significant supplement to knowledge representation learning (KRL). There are two main challenges for constructing knowledge representations from plain texts: (1) How to take full advantages of sequential contexts of entities in plain texts for KRL. (2) How to dynamically select those informative sentences of the corresponding entities for KRL. In this paper, we propose the Sequential Text-embodied Knowledge Representation Learning to build knowledge representations from multiple sentences. Given each reference sentence of an entity, we first utilize recurrent neural network with pooling or long short-term memory network to encode the semantic information of the sentence with respect to the entity. Then we further design an attention model to measure the informativeness of each sentence, and build text-based representations of entities. We evaluate our method on two tasks, including triple classification and link prediction. Experimental results demonstrate that our method outperforms other baselines on both tasks, which indicates that our method is capable of selecting informative sentences and encoding the textual information well into knowledge representations.
\end{abstract}

\section{Introduction}

Knowledge graphs (KGs), which provide significant well-structured information for modeling abstract concepts as well as concrete entities in real world, have attracted great attention in recent years. A typical knowledge graph usually arranges multi-relational data in the form of triple facts (\emph{head entity}, \texttt{relation}, \emph{tail entity}) that is abridged as $(h, r, t)$.

\begin{figure}[!htbp]
\centering
\includegraphics[width=0.99\columnwidth]{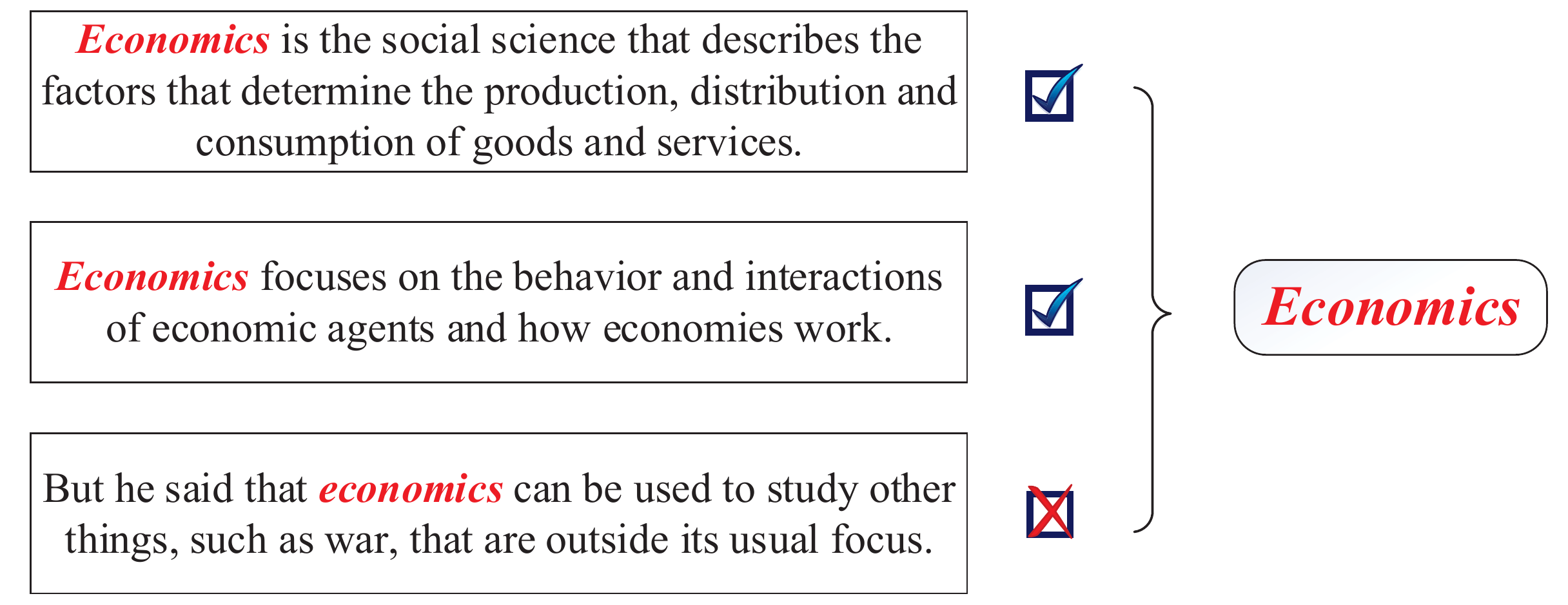}
\caption{An example of entity's reference sentences.}\label{fig. 1}
\end{figure}

There are large numbers of KGs like Freebase, YAGO and DBpedia that are widely utilized in nature language processing applications such as question answering and web search \cite{bollacker2008freebase}. However, applications for KGs are suffering from challenges of data sparsity and computational inefficiency with KG size increasing. To alleviate these problems, representation learning (RL) is proposed and widely used, significantly improving the capability of knowledge representations in knowledge inference, fusion and completion \cite{yang2014embedding,dong2014knowledge,neelakantan2015compositional}.

Many methods have introduced representation learning to KGs, projecting both entities and relations into a continuous low-dimensional vector space \cite{nickel2011three,jenatton2012latent,bordes2013translating}. Among existing methods, translation-based models, which interpret relations as translating operations between head and tail entities, are effective and efficient that possess the state-of-the-art performance.

Furthermore, rich external information is considered as supplement to triple facts that helps to represent knowledge graphs, and textual information has shown significant contribution to this goal. \cite{wang2014knowledge,zhong2015aligning} propose a joint model projecting both entities and words into the same vector space with alignment models. However, their models only consider bag-of-words assumption when modeling words in plain texts, neglecting rich textual information embedded in word order. \cite{xie2016representation} directly builds entity representations from entity descriptions, while their model is restricted by the completeness and quality of entity descriptions.

There are two main shortages in existing methods for constructing knowledge representations from multiple sentences in plain texts: (1) The bag-of-words assumption fails to encode explicit word order information into sentence representations. (2) Not all sentences containing entity names are reliable and suitable for explaining the corresponding entities. Fig. \ref{fig. 1} demonstrates an example of multiple reference sentences for the entity \emph{economics}. We can observe that the first two sentences talk about the definition and attributes of \emph{economics}, which could represent the entity well, while the third one provides rather confusing and meaningless information for understanding the meaning of \emph{economics}.

As shown in \cite{nagy1987learning}, humans learn meanings of new words from their contextual information. Inspired by this, we intend to infer the meaning of an entity from its reference sentences. To overcome the shortages mentioned above, we propose the Sequential Text-embodied Knowledge Representation Learning (STKRL). Specifically, in our method, we first utilize recurrent neural network (RNN) with pooling or long short-term memory network (LSTM) to build sentence-level representations of an entity from multiple sentences. Each sentence-level representation is considered as a candidate of the corresponding entity representation. Second, we combine these sentence-level representations to form the summary text-based representation, with the favor of attention which highlights more informative sentences. Finally, we follow the margin-based score function in translation-based methods as our objective for training.

To the best of our knowledge, our model is the first model which combines multi-instance learning with attention in knowledge representation learning with texts. We evaluate our models on triple classification and link prediction, and the significantly improved experimental results indicate our model is capable of representing knowledge graph better with textual information. Meanwhile, our model is of potential usefulness to definition extraction according to inspection towards the sentence attention component in our model.

\section{Related Work}

\subsection{Translation-based Methods}

Translation-based methods have achieved great successes for representation learning in knowledge graphs. TransE \cite{bordes2013translating} interprets relations as translating operations between head and tail entities, and projects both entities and relations into the same continuous low-dimensional vector space.
The energy function is defined as follows:
\begin{equation}
\begin{split}
E(h,r,t)=||\mathbf{h}+\mathbf{r}-\mathbf{t}||,
\end{split}
\end{equation}

which assumes that the tail embedding $\mathbf{t}$ should be in the neighborhood of $\mathbf{h}+\mathbf{r}$. TransE is straightforward and effective, while this simple translating operation may have issues when modeling 1-to-N, N-to-1 and N-to-N relations. Moreover, the translating operation only focuses on a single step, regardless of rich information located in long-distance relational paths. To address the first problem, TransH \cite{wang2014transH} models translating operations between entities on relation-specific hyperplanes. TransR \cite{lin2015learning} interprets entities and relations in different semantic spaces, and projects entities from entity space to relation space when learning the potential relationship between entities. TransD \cite{ji2015knowledge} proposes dynamic mapping matrix constructed by both entities and relations for multiple representations of entities. To extend the single-step translating operation, \cite{gu2015traversing,lin2015modeling} encode multiple-step relation paths into representation learning of knowledge graphs and achieve significant improvements.

\subsection{Representation Learning of Knowledge Graphs with Textual Information}

Multi-source information, especially textual information, is significant in representation learning of knowledge graphs, which has attracted great attention recently. It can provide useful information from different aspects, which helps modeling knowledge graphs. \cite{wang2014knowledge} encodes both entities and words into a joint low-dimensional vector space by alignment models using entity names as well as Wikipedia anchors. \cite{zhong2015aligning} extends the alignment model by considering entity descriptions. However, the methods of modeling plain texts in the both models above are rather simple, ignoring significant information encoded in word order. \cite{xie2016representation} proposes a new kind of representation for entities, which is directly constructed from entity descriptions using CNN and thus is capable of modeling new entities. However, their description-based representation is restricted by the completeness as well as quality of entity descriptions. To the best of our knowledge, our model is the first effort which learns knowledge representations from multiple sentences extracted from noisy plain texts with word order being considered.

\begin{figure*}[!htbp]
\centering
\includegraphics[width=0.9\textwidth]{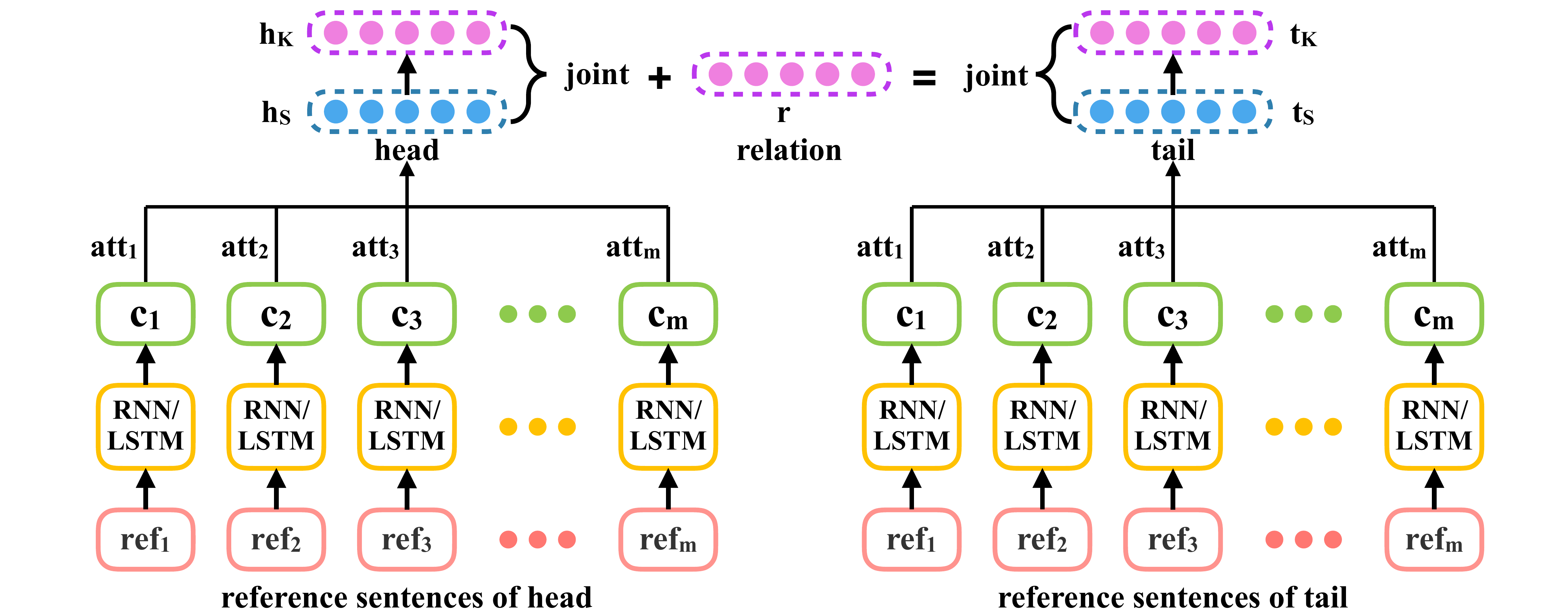}
\caption{Overall architecture of the STKRL model.}\label{fig. 2}
\end{figure*}

\subsection{Multi-instance Learning}

Multi-instance learning, which was originally proposed in \cite{dietterich1997solving}, arises in the tasks where a single object may possess multiple alternative examples or representations that describe it. Multi-instance learning aims to find out the reliability of examples in each object. \cite{bunescu2007learning} extends a relation extraction method using minimal supervision with multi-instance learning. \cite{zhou2012multi} proposes multi-instance multi-label learning on multiple tasks such as scene classification and text categorization. \cite{riedel2010modeling,hoffmann2011knowledge,surdeanu2012multi} adopt multi-instance learning in distant supervision for relation extraction. \cite{zeng2015distant} further combines multi-instance learning with convolutional neural network for relation extraction on distant supervision data. To fully utilize the rich textual information located in multiple sentences of each entity, we propose a sentence-level attention over multiple sentences to highlight more informative instances. To the best of our knowledge, multi-instance learning combined with attention-based model hasn't been adopted in representation learning of knowledge graphs.

\section{Methodology}

We first describe the notations used in this paper. For any triple $(h, r, t) \in T$, it consists of two entities $h, t \in E$ and a relation $r \in R$. $E$ stands for the set of entities while $R$ stands for the set of relations. $T$ is the training set of triples, and the embeddings of entities and relations take values in $\mathbb{R}^k$.

We have two representations for each entity. We set $\mathbf{h}_K$, $\mathbf{t}_K$ as the \textbf{structure-based representations} of head and tail entities, which are the same as those in previous knowledge models, and $\mathbf{h}_S$, $\mathbf{t}_S$ as the \textbf{text-based representations}, which are learned from plain texts by sentence encoders.

For each entity, we first scan through the corpus to extract all sentences which contain the corresponding entity name. These sentences are considered as the \textbf{reference sentences} of the corresponding entity.

\subsection{Overall Architecture}

First, we introduce the overall architecture of the STKRL model. Inspired by translation-based methods, we define the energy function as follows:
\begin{equation}
E(h,r,t) = E_{KK} + E_{SS} + E_{SK} + E_{KS},
\end{equation}

where $E_{KK}$ is the same energy function as TransE, $E_{SS}$, $E_{SK}$ and $E_{KS}$ are new-added parts determined by the two representations jointly. $E_{SS}=\Vert \mathbf{h}_S + {\bf r} - \mathbf{t}_S\Vert$ in which both head and tail are text-based representations learned from reference sentences. Similarly, we also have $E_{SK}=\Vert \mathbf{h}_S + {\bf r} - \mathbf{t}_K\Vert$ and $E_{KS}=\Vert \mathbf{h}_K + {\bf r} - \mathbf{t}_S\Vert$, jointly considering the two types of entity representations.

The overall architecture of the STKRL model is demonstrated in Fig. \ref{fig. 2}. First, The RNN/LSTM sentence encoder takes reference sentences as inputs and learns the sentence-level representations. Second, an attention method is implemented to select the top $m$ ($m$ is a hyper parameter) informative sentences to generate the text-based representation of an entity. The attention is based on the semantic similarity between the sentence-level representations and the corresponding structure-based representation. Finally, those representations will be learned jointly under the translation-based method.

\subsection{Word Representations}

In our framework, we consider the embeddings of each word token in reference sentences as inputs, and each entity name is also regarded as a word. Inspired by \cite{zeng2014relation}, the word representations consist of two parts, including word features and position features.

\subsubsection{Word Features}

The word features could be learned by negative sampling Skip-gram models, for these models could encode contextual information located in large corpus. The learned word embeddings are then directly considered as word features.

\subsubsection{Position Features}

Word order information is significant that it can help us to better understand sentences, and we also intend to highlight the position of the entity name in its reference sentences. Suppose each sentence is represented as a sequence $s = (x_1, x_2, \cdots , x_{n})$, where $x_i$ represents the $i$-th word. Given a sentence, the position feature of its entity name will be marked as $0$, and the positions of other words are marked according to the relevant integer distance towards the entity name. The left words have negative position values, while the right have positive position values. The position features will be marked as $-d$ or $d$ if their relevant distances are larger than $d$ ($d$ is a hyper parameter).

\subsection{Sentence Encoder}

We assume that the meaning of an entity could be extracted from its reference sentences. There are amounts of algorithms to represent sentence information with word order in consideration, such as recurrent neural network (RNN) and long short-term memory (LSTM). These models have been widely used in several natural language processing tasks such as machine translation. We adopt RNN with pooling and LSTM as sentence encoders to learn sentence-level representations of entities, intending to extract entity meanings from reference sentences.

\begin{figure}[!htbp]
\centering
\includegraphics[width=0.9\columnwidth]{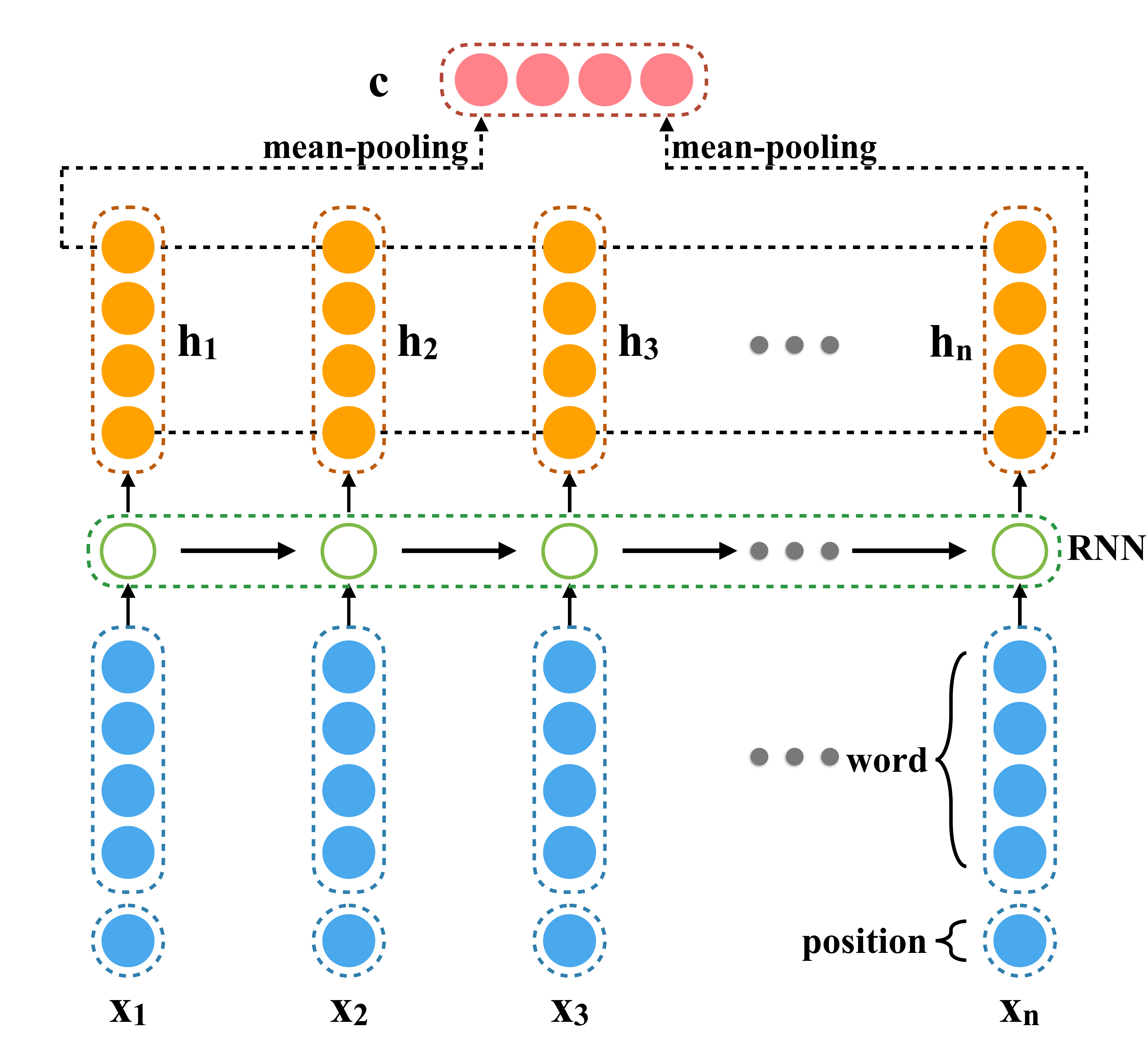}
\caption{Recurrent neural network with pooling.}\label{fig. 3}
\end{figure}

\subsubsection{Recurrent Neural Network}

Recurrent neural network takes a reference sentence as input. It works on a sequence and maintains a hidden state $h$ over time. At each time-step $t$, the hidden state vector ${\bf h}_{t}$ is updated as follows:
\begin{equation}
{\bf h}_{t} = \tanh(\mathbf{W}{\bf x}_t+\mathbf{U}{\bf h}_{t-1}+\mathbf{b}),
\end{equation}

in which transition function is an affine transformation followed by a non-linear function such as hyperbolic tangent. More specifically, RNN reads each word representation of the input sentence sequentially. While reading each word representation, the hidden state of the RNN changes according to Eq. (3). After finishing the whole sequence marked by an end-of-sequence symbol, we can obtain the final hidden state vector $\mathbf{h}_{n}$ as the output. $\mathbf{h}_{n}$ is then viewed as the sentence-level representation.

\subsubsection{RNN with Pooling}

Recurrent neural network is powerful and widely used in various fields, while it usually suffers from gradient vanishment. Therefore, it is difficult for the final hidden state of RNN to capture the early local information when the sentence is too long. \cite{collobert2011natural} proposes a mean-pooling approach to solve gradient vanishment to some degree. To leverage efficiency and effectiveness, we add a mean-pooling layer to encode the overall information of a sentence into its sentence-level representation $\mathbf{c}$. We have:
\begin{equation}
{\bf c} = \sum_{i=1,\cdots,n} \frac{{\bf h}_i}{n},
\end{equation}

in which all intermediate hidden states containing different local information should have contribution to the final sentence-level representation, and thus could be updated during back propagation. The structure is shown in Fig. \ref{fig. 3}.

\subsubsection{Long Short-Term Memory Network}

LSTM \cite{hochreiter1997long} is an enhanced neural network based on RNN, which could address the gradient vanishment when learning long-term dependencies. LSTM introduces memory cells that are able to preserve state over long periods of time. At each time step $t$, the LSTM unit is composed of: an input gate $i_t$, a forget gate $f_t$, an output gate $o_t$, a memory cell $c_t$ and a hidden state $h_t$. The entries of the gating vectors $\mathbf{i}_t$, $\mathbf{f}_t$ and $\mathbf{o}_t$ are in $[0,1]$. The LSTM transition equations are the following:
\begin{align}
\mathbf{i}_t & = \sigma(\mathbf{W}^{(i)}\mathbf{x}_t + \mathbf{U}^{(i)}\mathbf{h}_{t-1} + \mathbf{b}^{(i)}),
\\
\mathbf{f}_t & = \sigma(\mathbf{W}^{(f)}\mathbf{x}_t + \mathbf{U}^{(f)}\mathbf{h}_{t-1} + \mathbf{b}^{(f)}),
\end{align}
\begin{align}
\mathbf{o}_t & = \sigma(\mathbf{W}^{(o)}\mathbf{x}_t + \mathbf{U}^{(o)}\mathbf{h}_{t-1} + \mathbf{b}^{(o)}),
\\
\mathbf{u}_t & = \tanh(\mathbf{W}^{(u)}\mathbf{x}_t + \mathbf{U}^{(u)}\mathbf{h}_{t-1} + \mathbf{b}^{(u)}),
\\
\mathbf{c}_t & = \mathbf{i}_t \odot \mathbf{u}_t + \mathbf{f}_t \odot \mathbf{c}_{t-1},
\\
\mathbf{h}_t & = \mathbf{o}_t \odot \tanh(\mathbf{c}_t),
\end{align}

where $\mathbf{x}_t$ is the input at the current time step $t$, $\sigma$ denotes the logistic sigmoid function and $\odot$ denotes the elementwise multiplication.

\subsection{Attention over Reference Sentences}

We utilize sentence encoders to build sentence-level representations from reference sentences, next we want to integrate those sentence-level representations to the text-based representation for each entity. Simply considering the mean or max of sentence embeddings will suffer from noises or lost rich information. Instead of taking the mean/max sentence embeddings, we propose an attention method to automatically select sentences which can explicitly explain the meaning of the entity. The attention-based model is powerful and has been widely applied to machine translation \cite{bahdanau2014neural}, abstractive sentence summarization \cite{rush2015neural} and speech recognition \cite{chorowski2014end}. It can automatically highlight more informative instances from multiple candidates.

We implement an attention-based multi-instance learning method to select top-$m$ informative reference sentences for the corresponding entity from all candidates. More specifically, each entity $e$ has a structure-based representation $\mathbf{e}_K$. For a sentence-level representation $\mathbf{c}$ which belongs to entity $e$, the attention between $\mathbf{c}$ and $\mathbf{e}_K$ is as follows:
\begin{equation}
att(\mathbf{c},\mathbf{e}_K) = \frac{{\bf c} \cdot {\bf e}_K}{\Vert {\bf c}\Vert\cdot\Vert {\bf e}_K\Vert}.
\end{equation}

Reference sentences with higher $att({\bf c},{\bf e}_K)$ are expected to better represent their corresponding entity information more explicitly. Finally, we pick the top-$m$ reference sentences, and the text-based representation $\mathbf{s}$ is as follows:
\begin{equation}
{\bf s} = \sum_{i=1}^{m} \frac{att({\bf c}_i,{\bf e}_K)\cdot {\bf c}_i}{\sum_{i=1,\cdots,m} att({\bf c}_i,{\bf e}_K)}.
\end{equation}

\subsection{Objective Formalization}

We utilize a margin-based score function as our training objective. The overall score function is defined as follows:
\begin{multline}
L = \sum_{(h,r,t)\in T} \sum_{(h',r,t')\in T'} \max(\gamma+E(h,r,t)
\\
-E(h',r,t'),0),
\end{multline}

where $\gamma$ is a margin hyper parameter. $E(h,r,t)$ is the energy function stated above, which can be either $L_1$ or $L_2$-norm. $T'$ is the negative sampling set of $T$, which is defined as
\begin{multline}
T' = \{(h',r,t) \mid h'\in E\} \cup \{(h,r,t') \mid t'\in E\},
\end{multline}

in which the head and tail entities are randomly replaced by another entity. Note that for each entity, there are two types of entity representations, including the text-based representation and the structure-based representation. Hence we can learn these two types of entity representations simultaneously into the same vector space.

The energy function (2) is used to learn structure-based and text-based representations into the same vector space, and the single $E_KK$ item in (2) is the energy function of TransE. Thus, as mentioned above, we can train our model using the objective formalization (13).

\subsection{Model Initialization and Optimization}
{}
The STKRL model can be initialized either randomly or with pre-trained TransE embeddings. The word representations are pre-trained through Word2Vec with Wikipedia corpus. The optimization is a standard back propagation using mini-batch stochastic gradient descent (SGD). For efficiency, we also use GPU to accelerate the training process. Note that there is \textbf{only one RNN/LSTM}, which means that all RNN/LSTM in the figure shares the same parameters, and this RNN/LSTM is trained to encode reference sentences for all entities.

Our model consists of two parts: (1) text-based RNN/LSTM with attention for encoding sentences, and (2) structure-based TransE \cite{bordes2013translating}for encoding knowledge. We train the two parts simultaneously. The training process is described as follows:
\begin{itemize}
\item Given a triple, learn structure-based and text-based knowledge representations using extended TransE with negative sampling. Note that, there are four terms for each triple, ($h_K$, $r$, $t_K$), ($h_S$, $r$, $t_K$), ($h_K$, $r$, $t_S$) and ($h_S$, $r$, $t_S$), with four terms of negative samples respectively.
\item Utilize RNN/LSTM to encode all reference sentences of head and tail entities.
\item Attention scheme is applied to dynamically select top-m informative sentences of head and tail from all encoded sentences and construct text-based representations respectively.
\item For selected $m$ sentences, the structure-based representation of head, $h_K$, (respectively, tail, $t_K$) is used to compute training error and optimize RNN/LSTM with back-propagation.
\end{itemize}

\section{Experiment}

\subsection{Datasets}

We adopt FB15K to evaluate our models on triple classification and link predication in this paper. FB15K is a subset of Freebase that has 14,951 entities and 1,345 relations. For each entity, we use Wikipedia as the corpus to extract reference sentences. Note that the Wikipedia article about the entity itself has the highest priority to provide reference sentences. Each entity has approximately 40 sentences. The statistics of the FB15K dataset are listed in Table \ref{Statistics}.

\begin{table}[ht]
\small
\caption{\label{Statistics}Statistics of dataset.}
\begin{center}
\begin{tabular}{p{27pt}<{\centering}|p{21pt}<{\centering}|p{24pt}<{\centering}|p{29pt}<{\centering}|p{26pt}<{\centering}|p{26pt}<{\centering}}
\hline 
Dataset & \#Rel & \#Ent & \#Train & \#Valid & \#Test \\ 
\hline
FB15K & 1,345 & 14,951 & 483,142 & 50,000 & 59,071 \\
\hline
\end{tabular}
\end{center}
\end{table}

\subsection{Parameter Settings}

The detailed parameters will be released upon acceptance.

\begin{table*}[ht]
\small
\caption{\label{link_2} Evaluation results on link prediction with different relation categories.}
\begin{center}
\begin{tabular}{@{}c|c|c|c|c|c|c|c|c@{}}
\hline Tasks & \multicolumn{4}{@{}c|}{Predicting Head (Hits@10)} & \multicolumn{4}{@{}c@{}}{Predicting Tail (Hits@10)} \\ \hline
Relation Category & 1-to-1 & 1-to-N & N-to-1 & N-to-N & 1-to-1 & 1-to-N & N-to-1 & N-to-N \\  \hline
TransE & 78.2 & 87.6 & 42.1 & 70.0 & 79.3 & 48.9 & 87.9 & 71.2 \\
Wang's & 84.6 & 89.4 & 49.3 & 72.1 & 81.8 & 66.4 & 92.3 & 76.7 \\
\hline
STKRL (RNN w/o ATT) & 81.9 & 87.2 & 45.7 & 71.8 & 80.8 & 52.6 & 91.2 & 73.2 \\
STKRL (RNN+ATT) & 84.5 & 89.9 & 48.9 & 73.0 & 81.6 & 64.5 & 92.3 & 76.9 \\
STKRL (RNN+P+ATT) & 87.0 & 90.9 & 52.0 & \textbf{77.6} & 85.8 & 67.1 & 93.9 & 82.2 \\
STKRL (LSTM+ATT) & \textbf{88.9} & \textbf{91.6} & \textbf{53.2} & 77.0 & \textbf{85.9} & \textbf{68.3} & \textbf{94.2} & \textbf{83.6} \\
\hline
\end{tabular}
\end{center}
\end{table*}

For STKRL, we implement 4 sentence encoders for evaluation. ``RNN w/o ATT'' represents selecting RNN as sentence encoder and using the mean vector of sentence-level representations instead of attention, while ``RNN+ATT'', ``RNN+P+ATT'' and ``LSTM+ATT'' represent selecting the corresponding sentence encoders of RNN, RNN+pooling and LSTM with the help of attention.

We implement TransE and jointly (name) model in \cite{wang2014knowledge} as baselines, following their experimental settings. For fair comparisons, all baselines have the same dimension of entities and relations.

\subsection{Triple Classification}

The task of triple classification aims to test whether a triple $(h, r, t)$ is true or false, which could be viewed as a binary classification test.

\subsubsection{Evaluation Protocol}

Since each entity has two types of representations, a triple $(h,r,t)$ has four representations, $(h_{K},r,t_{K})$, $(h_{S},r,t_{S})$, $(h_{S},r,t_{K})$ and $(h_{K},r,t_{S})$. Following the similar protocol in \cite{socher2013reasoning}, for each triple representation, we construct a negative example by randomly replace the head or tail entity with another entity. The new entity's representation should be the same type as the replaced one (e.g. $(h_{K},r,t_{K})$ should be replaced with $(h'_{K},r,t_{K})$ or $(h_{K},r,t'_{K})$). And we also ensure that the number of true triples is equal to false ones.

The evaluation strategy is described as follows: if the dissimilarity of a testing triple $(h, r, t)$ is below the relation-specific threshold $\delta_r$, it is predicted to be positive, otherwise negative. The threshold $\delta_r$ can be determined via maximizing the classification accuracy on the validation set.

\subsubsection{Results}

The results of triple classification are shown in Table \ref{triple}. From the results, we observe that: (1) Our attention-based models significantly outperform all baselines, which indicates the capability of our methods in modeling knowledge representations. (2) STKRL (RNN w/o ATT) outperforms TransE, which implies the significance of textual information. Moreover, the attention-based STKRL models outperform Wang's method which also encodes textual information into knowledge representations. It confirms that our sentence encoders can better understand the meaning of sentences, and thus get better performances. (3) Attention is the key component in our model. It can automatically select the more informative reference sentences to represent an entity, alleviating the noises caused by low-quality reference sentences. (4) STKRL (LSTM+ATT) achieves the best performance, successfully capturing the rich information in long-distance dependencies. Besides, STKRL (RNN+P+ATT) outperforms STKRL (RNN+ATT), which also demonstrates the advantages of pooling strategy.

\begin{table}[!h]
\small
\center
\caption{\label{triple} Evaluation results on triple classification.}
\begin{tabular}{@{}c@{}c@{}}
\hline \bf Method & FB15K  \\ \hline
TransE & 77.1  \\
Wang's & 84.2  \\
\hline
STKRL (RNN w/o ATT) & 79.2 \\
STKRL (RNN+ATT) & 85.8 \\
STKRL (RNN+P+ATT) & 87.9 \\
STKRL (LSTM+ATT) & {\bf 88.6} \\
\hline
\end{tabular}
\end{table}

\subsection{Link Prediction}

The task of link prediction aims to complete a triple fact $(h, r, t)$ when $h$ or $t$ is missing.

\begin{table*}[ht]
\small
\caption{\label{economics} The reference sentences of \emph{economics} ranked by attention.}
\begin{center}
\begin{tabular}{@{}c|p{400pt}@{}}
\hline \bf Rank & \bf Sentence  \\ \hline
1 & \emph{\textbf{Economics}} is the social science that describes the factors that determine the production, distribution and consumption of goods and services.  \\
2 & \emph{\textbf{Economics}} focuses on the behavior and interactions of economic agents and how economies work.  \\
10 & The ultimate goal of \emph{\textbf{economics}} is to improve the living conditions of people in their everyday life. \\
44 & There are a variety of modern definitions of \emph{\textbf{economics}}. \\
\hline
\end{tabular}
\end{center}
\end{table*}

\begin{table*}[ht]
\small
\caption{\label{sentences} The rank No.1 reference sentences of different entities.}
\begin{center}
\begin{tabular}{@{}c|p{350pt}@{}}
\hline \bf Entity & \bf Rank No.1 Sentence  \\ \hline
\emph{Productivity} & \emph{\textbf{Productivity}} is the ratio of output to inputs in production.  \\
\emph{February} & {\bf \emph{February}} is the second month of the year in the Julian and Gregorian calendars. \\
\emph{Food} & {\bf \emph{Food}} is any substance consumed to provide nutritional support for the body. \\
\emph{Travis County} & Austin is the capital of Texas and the seat of {\bf \emph{Travis County}}.  \\
\hline
\end{tabular}
\end{center}
\end{table*}

\subsubsection{Evaluation Protocol}

For each test triple, the head is replaced by each of the entities of the entity set in turn. Dissimilarities of those corrupted triples are first computed by the models and then sorted by ascending order. The rank of the correct entity is finally stored. This whole procedure is repeated while removing the tail instead of the head, and the same rule applies to the tail. Note that in Wang's and our model, each entity has two representations. As a result, the predicted rank for a certain entity is the mean of two representations. However, as for Hits@10 test, either of two representations appearing at top 10 could be viewed as a successful hit.

\subsubsection{Results}

From Table \ref{link_1}, we can observe that: (1) Our model significantly outperforms all baselines in both Mean Rank and Hits@10, which demonstrates the effectiveness and robustness of our model. (2) STKRL (RNN+P+ATT) and STKRL (LSTM+ATT) significantly outperform Wang's method. It is because that we utilize sentence encoders to model textual information with the help of word orders, instead of simply considering separate words used in Wang's method. Moreover, the attention-based STKRL models could pay more attention to those informative reference sentences when constructing text-based representations. (3) STKRL (LSTM+ATT) achieves the best performances, which indicates the power in utilizing better sentence encoders.

\begin{table}[ht]
\small
\caption{\label{link_1} Evaluation results on link prediction.}
\begin{center}
\begin{tabular}{@{}c|c@{}c|c@{}c@{}}
\hline \bf Metric & Mean & Rank & Hits@ & 10($\%$) \\
& Raw & Filter & Raw & Filter \\  \hline
TransE & 215 & 112 & 46.5 & 68.7 \\
Wang's & 184 & 67 & 50.9 & 78.2 \\
\hline
STKRL (RNN w/o ATT) & 207 & 96 & 47.4 & 71.9 \\
STKRL (RNN+ATT) & 195 & 73 & 50.9 & 78.4 \\
STKRL (RNN+P+ATT) & 188 & 56 & {\bf 52.2} & 80.6 \\
STKRL (LSTM+ATT) & {\bf 182} & {\bf 51} & 52.1 & {\bf 81.2} \\
\hline
\end{tabular}
\end{center}
\end{table}

Table \ref{link_2} demonstrates the link prediction results with different categories of relations. Relations are split into 4 categories following the same settings in \cite{bordes2013translating}. From Table \ref{link_2} we can observe that: (1) the STKRL model achieves great improvements consistently on all categories of relations. (2) The improvements locate more in N-to-N relations. It indicates that textual information is important when modeling complex relations, and our STKRL models could well capture the useful textual information for better knowledge representations.

\subsection{Case Study}

In this section, we give two cases and analysis to prove that we can successfully select informative reference sentences to represent entities.

\subsubsection{Effectiveness of Attention}

To show that the attention has reasonably ranked reference sentences, we give 4 examples of the entity \emph{economics} with their ranks. The results in Table \ref{economics} indicate that STKRL can successfully select the more informative sentences of the corresponding entity. Those top-ranked sentences will usually be the definitions or descriptions of entities. Besides, the low-ranked sentences are usually less relevant to the targeted entity, which also indicates the capability of attention in filtering noises.

\subsubsection{Definition Extraction}

Definition extraction is an important task in text mining \cite{espinosa2015weakly}. We want to demonstrate that the top-ranked reference sentences selected by our model are usually definitions of the corresponding entities.

By randomly selecting 100 entities from the entity set $E$, we find that 61 entities' No.1 reference sentences are entity definitions based on manual annotation. In Table \ref{sentences} we give 4 No.1 reference sentence examples. The first three sentences are exactly entity definitions, while the last is not. However, all sentences are still informative to learn the meanings of entities. The results demonstrate that our model has rational selectivity to extract reference sentences and is of potential usefulness to definition extraction.

\section{Conclusion}

In this paper, we propose the STKRL model, a novel model for knowledge representation learning, which jointly considers triple facts and sequential text information. We also explore how to extract informative sentences from lots of candidates with attention-based method. We evaluate our models on two tasks, triple classification and link prediction, and also give some examples to prove the ability to efficiently extract useful information. Experimental results show that our models achieve significant improvements compared to other baselines. The code and dataset will be released upon acceptance.

We will explore the following further work: (1) We assume that for entity representations, there are naturally three types of representations including word, sentence and knowledge representations. We mainly use knowledge and sentence representations in the STKRL model, and we will explore to integrate all three representations in future. (2) The STKRL model can extract definition sentences of entities as by-products. In future, we will further explore this issue by designing a more sophisticated method for definition extraction.

\bibliography{emnlp2016}
\bibliographystyle{emnlp2016}

\end{document}